# When the harness decides success: state that cannot be re-observed

**Raimondas Lencevicius**
Microsoft
rlencevicius@microsoft.com

October 2026

**Abstract**

Agent harnesses decide what a language model sees across turns: they drop old turns, summarize them, or replay the model's own earlier reasoning. We ask when these choices change whether an agent succeeds, not only what it costs, and report three findings. First, in a pre-registered synthetic probe, truncating the context window costs a low-cost reasoning model nothing in success when a needed fact can be looked up again (it looks the fact up instead), but drops its success to chance when the fact cannot be re-observed: the two penalties differ by 92.5 percentage points (90% CI [89.5, 95.5]). A pre-registered replication on one model from a second provider showed the same contrast. Once the fact was lost, that model chose no answer rather than guessing. Second, on ARC-AGI-3, where the benchmark's Standard harness trails its Provider Adapter harness by a wide margin, we decompose that gap one component at a time, for one model on four games. Summarizing old turns instead of dropping them, while keeping the Standard harness's invitation to take notes, recovers 86% of a 63-point gap without replaying the model's reasoning. The same harness without the invitation recovers 43%, and adding the provider's native replay of the model's reasoning recovers the remaining 57%. These are point estimates from one counted run per configuration and game, and the replay share depends on a single run. Third, in the benchmark's published runs and in ours, the Adapter's advantage in the probability of clearing a level it has reached is larger at deep levels than at early ones. This largely restates where runs stopped and does not separate a growing advantage from a constant one.

## 1 Introduction

A model's context is finite, so every agent harness has a policy for what to keep when an episode outgrows it. The simplest drops the oldest turns. Others replace them with a summary, ask the model to keep notes, or, where the provider supports it, replay the model's own reasoning items from earlier turns. These policies are often chosen for cost and latency, on the implicit assumption that a capable model can recover whatever it loses. That assumption fails in a specific case: when the information the agent needs was *derived* during the episode, or was shown once and cannot be shown again. Then a lost fact is lost for good, and the harness, as much as the model, decides whether the agent succeeds.

We test this in two settings.

**A controlled probe.** PersistGym-lite generates episodes in which a fact needed at a later step is either shown or derived from clues, and is either recoverable (a lookup re-shows it) or not. Holding the model fixed, we compare a full-history harness with a truncated window, with and without a *note-taking instruction*: one system-prompt sentence that invites the model to repeat, in each visible reply, anything it wants to carry forward. Our first hypothesis (H1) is that truncation costs success when the fact cannot be re-observed and costs only effort when it can. A control (H3) checks that multi-turn structure itself costs at most a few points of success when the fact survives. H2, a test of reasoning replay on a second model from the same provider, was stopped (§8). A later pre-registered amendment adds a replication of H1 and H3, outside the confirmatory tests, on one model from a second provider (§3–4).

**A public benchmark.** ARC-AGI-3 (ARC Prize Foundation 2026b) is a benchmark of interactive games in which an agent must discover the rules as it plays. Its own harness (ARC Prize Foundation 2026c) has two configurations. *Standard* keeps a plain-text history of the observations and the model's visible replies, without its reasoning, and drops the oldest turns when the context fills. *Provider Adapter* uses each provider's native conversation format, including the model's reasoning items, and summarizes old turns instead. They give the same model very different scores (ARC Prize Foundation 2026d; Kamradt 2026). We decompose that gap for one model on four games, changing one component at a time (H4):

- the *loss policy*: dropping the oldest turns, or replacing them with a model-written summary;
- the *note-taking instruction* (above): present in the Standard harness and absent from the Adapter, it keeps a fact in view after the turns that produced it are gone, as long as the model restates it in later replies;
- *native reasoning replay*: with each new request, the harness sends back the model's reasoning items from earlier turns in the provider's own format (for the model we test, its thoughts with their thought signatures (Google 2026b)), whereas the Standard harness sends back only the visible reply text.

We then ask where in a game the gap arises (H5).

Two of the mechanisms behind the ARC gap have been named before. One provider attributes its model's gain to retained reasoning and compaction together, reporting only their combined effect (Bigio and Sanders 2026), and VISTA adds components one at a time to the benchmark's official implementation, among them compaction in place of a sliding window and notes files, on the way to its own visual harness (Han et al. 2026). Neither reports native reasoning replay as a separate step, and neither decomposes the benchmark's own Standard-to-Adapter gap. We do, for a model from another provider, with the Standard harness's note-taking instruction as a component of its own, and we pair the decomposition with a controlled probe.

Following pre-registration practice (Nosek et al. 2018), the hypotheses, predictions, thresholds and primary analyses were frozen in the project repository (`PREREG.md` v1.0, 2026-10-01) before the first paid run. Later amendments cut the scope, added secondary readings, changed one ARC setting (the summary budget of two configurations) and added the replication, several after some results had been seen. Each is recorded with its date and reason, and §8 lists them. Analyses marked exploratory were not registered. **Our contributions:**

1. a probe that separates "the fact was lost" from "the fact can be re-observed", with its pre-registered result and a pre-registered replication on one model from a second provider (§3–4);
2. a component-by-component decomposition of ARC-AGI-3's Standard-to-Adapter gap for one model, at one run per configuration and game (§5);
3. a per-level description of how the two harnesses' clear rates differ, on the benchmark's published runs and ours (§6);
4. an open record: pre-registration, amendments, decisions log, code and a $750 cost ledger (§10).

## 2 Related work

**Context management for agents.** LOCA-bench grows an agent's context while keeping task semantics fixed, and finds that performance falls as the environment state grows, while advanced context-management techniques substantially raise success (Zeng et al. 2026). For coding agents, an empirical study of harness design compares five context-management strategies across window sizes and finds that management matters more as windows shrink, mainly by preventing overflow, and that making elided content recoverable added no accuracy (Fan et al. 2026). Eviction can also cost success outright: on ALFWorld, naive eviction lowered pooled success from 56.7% with full history to 22.0%, and keeping or re-surfacing only the agent's plan did not close the gap (Mehta and Datta 2026). Neither study separates information the agent can look up again from information it cannot. Our probe isolates that case: a fact that, once dropped, cannot be observed again. For an agent's own earlier reasoning, M. Wang et al. (2026) argue that it matters less once the state it derived has been moved into code, files, tool outputs or environment feedback, which fits our framing: what must be carried is derived state that cannot be read again. Ledger turns interaction history into explicit execution state for coding agents and reports higher resolution rates at lower cost on SWE-bench Verified (Z. Wang et al. 2026). We study when such state is lost, not how to build it.

**Harnesses on ARC-AGI-3.** Several systems build their own agent around ARC-AGI-3's games: programmatic search over a complete, structured interaction log (PRO-LONG, Fox et al. 2026), a persistent Python REPL plus a layer that carries histories, memories and skills across trajectories (Prime Agent, Karten et al. 2026), executable world models built and repaired during play (Tycho, Lehmann et al. 2026), and a visual harness with a lossless memory of past observations (VISTA, Han et al. 2026). Wu et al. (2026) trace the beliefs, rules and plans a coding agent commits to its notes and scripts while playing the benchmark. In the benchmark's Kaggle competition (Chollet et al. 2026), which runs open-weight models offline, all three Milestone 2 prize winners build on the open-sourced Duck harness (Bessis et al. 2026; Kang 2026). The Duck is configured to keep each turn's reasoning in the history it sends back to the model (one competitor's audit reports that in the stock code this reasoning does not actually reach the prompt (Canivel 2026)), carries a structured world-model note (world, goal and action models, open questions) from turn to turn, and drops the oldest turns when its context fills (Bessis et al. 2026). A public competitor log reports experiments with compaction plus retained reasoning and with a structured ledger and leaves both unresolved (Canivel 2026), and a competition post that added one layer at a time to a Duck-based harness found almost every layer at or below its baseline, within run-to-run noise (Makarov 2026).

The gap itself is documented. ARC Prize reports the Standard and Provider Adapter harnesses across reasoning efforts for one model (Kamradt 2026), and one provider attributes its own model's gain to two settings, retained reasoning and compaction in place of truncation, reporting only their combined effect (Bigio and Sanders 2026). We do not build a new agent. We take the benchmark's own harness (ARC Prize Foundation 2026c) and decompose the published gap between its Standard and Provider Adapter configurations into its components, for one API model.

## 3 PersistGym-lite

PersistGym-lite generates short agent episodes built to isolate one question: when an agent needs a fact late in an episode, is that fact still available to it, and can it get the fact back? Real agent tasks mix this with skill, search and luck. In PersistGym-lite we control where the fact appears, how the agent learns it (shown directly, or derived from clues it must combine), whether it can be shown again, and how far back it lies relative to what a truncated window keeps. Everything else in an episode is routine play: it puts distance between the reveal and the need, so that a truncated window can lose the fact, and it checks basic competence. Every episode is played under every harness, so harness effects are compared on identical episodes.

**Episodes.** Each episode is a sequence of steps the agent must act on. *Routine* steps depend only on the current observation. A *reveal* step shows the target fact (observed), or one of several clues and, at some step, the rule that combines them (derived). A later *need* step asks for an action whose parameter is the fact, chosen from K = 10 options, so chance is 10%. In the recoverable variant a *lookup* action re-shows the fact (or its clues) at the cost of a step.

**Task cells.** Observed or derived fact × recoverable or not × near or far (whether the evidence lies inside or outside a truncated window at the need step): 8 cells. A generator draws T = 200 templates, and each template yields one episode per cell.

**Harness cells.** The suffix -0 marks cells that do not replay the model's reasoning. F-0 keeps the full history. W-0 keeps a window of the most recent turns, dropping the oldest, within W = 24,000 tokens. W-n-0 adds the note-taking instruction to W-0: the system-prompt sentence "Include any context you want to carry forward in your reply, along with the action you want to take.", the same sentence ARC-AGI-3's Standard harness uses (§5). Σ-n-0 replaces dropped turns with a model-written summary and keeps the notes instruction (exploratory). A single-prompt control flattens an F-0 episode into one message, with the evidence at the same offset from the question (H3). A restated-at-need control (F-0 with the fact repeated at the need step) gives the memory-free ceiling. On gpt-6-luna it scored 100% in every cell. Before gpt-6-luna's sweep, the registered validity gate passed: with the evidence restated at the need step, 160 of 160 validity episodes were answered correctly, and with it withheld, 19 of 160 (11.9%, against a gate of 20%). Per cell, reported descriptively, the withheld run scored 0 to 6 of 20, the most in the derived, not-recoverable, far cell. The programmatic oracle scored 100%, and no memory-free heuristic scored above 10%.

**Model and protocol.** gpt-6-luna (OpenAI 2026b), a low-cost reasoning model, at medium reasoning effort on the provider's Flex tier, a half-price, slower processing tier (OpenAI 2026a) (every request was served on it), run on all 200 templates (repeat 0) and on 20 of them again (repeat 1). T = 200 is the cap set by amendment R6. The registered power rule would have set 290, and at T = 200 the simulated power at the worst Holm step was 0.65 for H1 and 0.78 for H3 (primary scenario), below the registered target of 0.80. Hypotheses, tests, thresholds and the analysis code were fixed before any result was read, except the changes in §8. A hypothesis is supported only if a one-sided bootstrap test (a template-cluster bootstrap with 10,000 resamples (Efron and Tibshirani 1993)) rejects after Holm correction across H1 and H3 (Holm 1979), and a model check, a logistic model with template fixed effects (Seabold and Perktold 2010), agrees. Where an outcome cell has no variation, the model check uses Firth's penalized fit (Firth 1993; Heinze and Schemper 2002), which we implement with SciPy (Virtanen et al. 2020).

**Replication on a second provider's model.** Amendment R8 (2026-10-18) registered a replication of H1 and H3 on claude-haiku-5-5 (Anthropic 2026), a small, low-cost model from a second provider, at medium effort with adaptive thinking. As in gpt-6-luna's -0 cells, the model sees none of its earlier reasoning, here because none is sent: each earlier turn goes back as its visible text only (an empty or refused turn as a short placeholder). It played the same 200 templates, episode for episode, in F-0, W-0 and both controls, with 20 templates again at repeat 1. A registered cost rule, applied before any main episode, dropped W-n-0. Before the sweep it passed the same validity gate: 160 of 160 restated and 0 of 160 withheld. Its restated-at-need control scored 198 to 200 of 200 per cell, the misses mostly refusals (§4). Its tests use the same predictions, bootstrap and model checks, but they are reported with unadjusted one-sided p-values outside the Holm family and carry no confirmatory label: the replication neither confirms nor refutes H1 or H3. At T = 200, a true $\Delta_{H1}$ of 10 points and one-sided 0.05, its simulated power was 0.91 for H1 and 0.86 for H3 under the primary assumptions (0.53 and 0.57 under pessimistic ones). This describes the design, not these runs. R8 was written before any output from this model existed, but with gpt-6-luna's results known.

## 4 PersistGym-lite results

**H1: truncation costs success only when the fact cannot be re-observed.** With full history (F-0), gpt-6-luna answered the need step correctly in every observed-fact, far-distance episode, recoverable or not (200/200 each). With a truncated window (W-0), it scored 7.5% (15/200) when the fact could not be looked up again, close to the 10% chance level, and 100% when it could (Figure 1). There it looked the fact up 201 times in 200 episodes. The difference between the two penalties is $\Delta_{H1}$ = +92.5 percentage points (90% CI [89.5, 95.5]). The pre-registered one-sided test rejects $\Delta_{H1} \leq 0$ after Holm correction, and the model check agrees: a Firth-penalized logistic fit gives a window-by-recoverability coefficient of +7.09, and a Bayesian logistic model with template random intercepts (Capretto et al. 2022; Abril-Pla et al. 2023) gives a posterior median of 5.25 (90% interval [2.86, 7.52]). H1 is supported. All four registered H1 predictions also held: the manipulation check (a penalty of 92.5 points, against a bar of 25), recoverable success (a penalty of 0, within 5), lookups (none under full history, 201 under the window) and need-phase tokens (below).

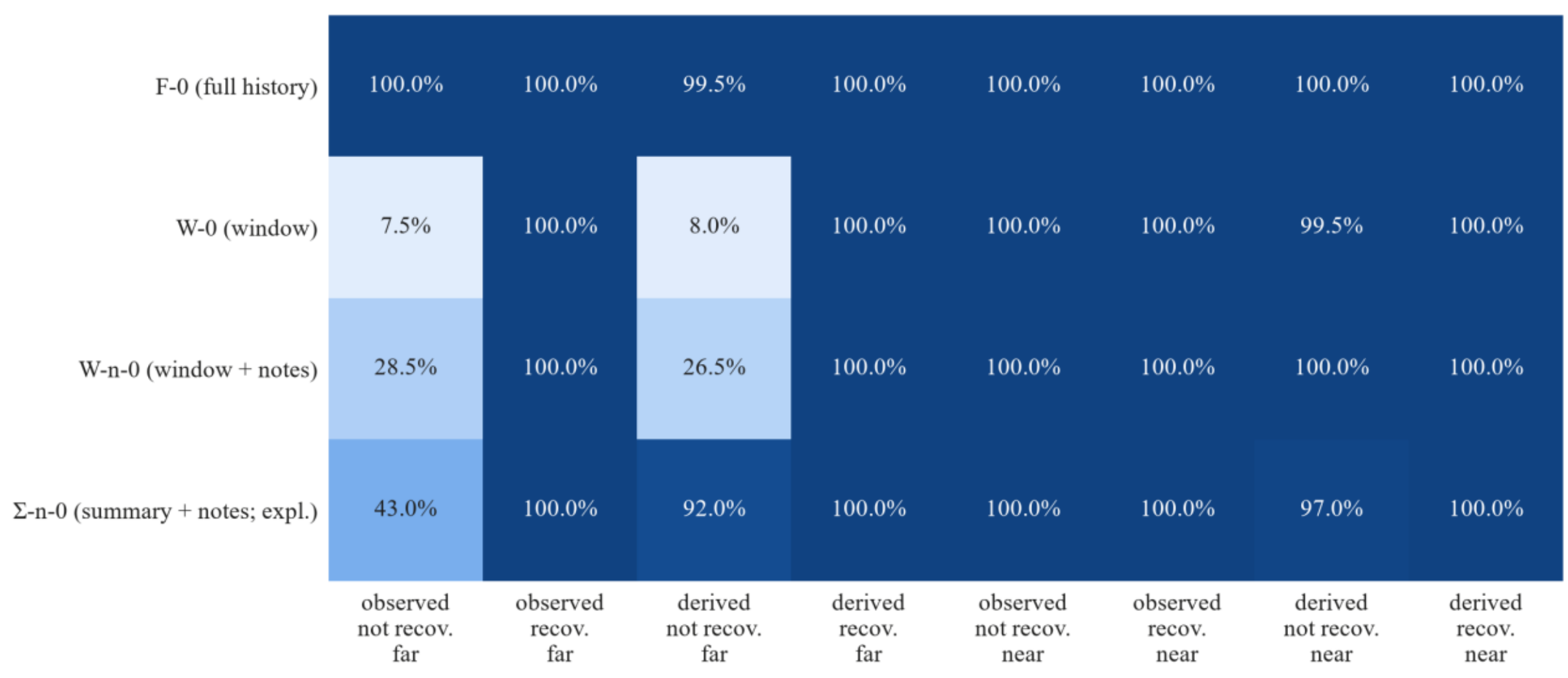

Figure 1: Success rate by harness cell and task cell (gpt-6-luna, 200 templates, repeat 0). Only the not-recoverable far cells separate the harnesses.

The window's price in the recoverable case is effort: it looked the fact up again, and its tokens on the need step and its lookups were 1.31 times the full-history harness's (45,875 against 35,088 per episode). Measured against near episodes, where the window lost nothing and its shorter context made those calls cheaper, the rise is 1.99-fold (the registered ratio, 90% CI [1.97, 2.01], against a registered bar of 1.30). The window also cost more in dollars, but not because of the lookups: it cost 2.9 to 4.0 times as much as full history in every task cell, near cells included (4.0 times in this cell, 95% CI [3.95, 4.03]), while using 0.84 to 0.90 times the tokens. Dropping turns breaks prompt caching: in the main runs, 94% of the full-history harness's input tokens were served from cache, against 48% for the window, and this model's cached input costs a tenth of its uncached price.

**H3: multi-turn structure costs at most a few points when the fact survives.** Flattening the same episode into one prompt gave 400/400 against 399/400 for the multi-turn conversation: a difference of +0.25 points (90% CI [0.0, 0.75]), well inside the registered 5-point bound. H3 is supported. Both arms are at ceiling, so this bounds the cost for this model and task, but it does not show that the cost is zero. With only one template showing any failure, the model check reproduces the raw difference and adds no independent evidence.

**Registered descriptive readings and exploratory results (not tests).** The note-taking instruction rescued part of the loss: the window with notes (W-n-0) scored 28.5% on the same cell, and in the episodes where its notes held the fact (21.5%), the model used it every time. Exploratory: in the not-recoverable far cells, the summary harness with notes (Σ-n-0) succeeded on 92% of derived-fact episodes against 43% of observed-fact ones. Its summary or notes still held the fact in 92% and 38.5% of those episodes. Also exploratory: in the observed, not-recoverable far cell, the three harnesses that could lose the fact spent 3.7 to 4.3 times as many reasoning tokens as full history (W-0 3.7, Σ-n-0 3.8, W-n-0 4.3).

**Disclosure.** Both H1 and H3 first read "inconclusive" because of code limits found after the results were read: the Firth fit stopped three steps short of its tolerance, and H3's check refused to fit with only one varying template. We fixed both, and report the labels the analysis first produced alongside the corrected ones (§8).

**Replication on a second provider's model.** On claude-haiku-5-5 the pattern was the same (Table 1). With a truncated window it succeeded in 0 of 200 episodes when the fact could not be looked up again (it chose no option, as explained below), and in 200 of 200 when it could, looking the fact up once in every episode. With full history it scored 200 and 198 of 200, respectively. $\Delta_{H1}$ = +101.0 points (90% CI [100.0, 102.5]): the one-sided test rejects ($p < 0.0001$, unadjusted), the Firth model check agrees (a window-by-recoverability coefficient of +8.55), and all four H1 predictions held. In H3's cells both arms were again at ceiling, 399 of 400 for the single prompt against 400 of 400 for the conversation: $\Delta_{H3}$ = −0.25 points (90% CI [−0.75, 0.0]), within the 5-point bound. As on gpt-6-luna, this bounds the cost at these tasks without showing that it is zero, and says nothing about harder ones. Only one template varies (a refusal), so H3's model check reproduces the raw difference and adds no independent evidence.

Table 1: H1 and H3 on two models, one per provider. Rows 1–2 give successes out of 200 observed-fact far episodes per cell. gpt-6-luna's tests are confirmatory, and claude-haiku-5-5's are a replication outside the Holm family. Each column gives one model's harness effects. Differences between the columns are described, not tested.

| | gpt-6-luna | claude-haiku-5-5 |
|---|---:|---:|
| Full history (F-0): not recoverable / recoverable | 200 / 200 | 200 / 198 |
| Window (W-0): not recoverable / recoverable | 15 / 200 | 0 / 200 |
| $\Delta_{H1}$, points (90% CI) | +92.5 [89.5, 95.5] | +101.0 [100.0, 102.5] |
| H3 cells, of 400: single prompt / conversation | 400 / 399 | 399 / 400 |
| $\Delta_{H3}$, points (90% CI) | +0.25 [0.0, 0.75] | −0.25 [−0.75, 0.0] |
| Reading | supported (confirmatory, after the fixes above) | unadjusted tests reject, H1's check agrees, H3's is degenerate (replication) |

Three details matter for reading these numbers.

- **The larger $\Delta_{H1}$ is not a larger memory loss.** When the window had dropped the fact, claude-haiku-5-5 never chose one of the 10 answer options. This held in all 400 not-recoverable far episodes, with observed and with derived facts, and a reply without a choice scores 0. gpt-6-luna chose an option in 387 of 400 and was right in 31 (7.75%), near chance. So 7.5 of the 8.5-point difference in $\Delta_{H1}$ is this difference in answering, and the remaining point comes from refusals (next item). The difference is described, not tested.
- **Refusals.** claude-haiku-5-5 refused some steps (refusal category "bio") in 1.1% to 1.6% of episodes in each multi-turn cell and in 1 of 800 single prompts. 13 of the 14 templates involved use the generator's biobank setting (a laboratory sample-tracking console). This is a property of these synthetic scenarios, not a general claim about the model. Refusals score 0 under the amendment. Two need-step refusals under full history make the window's penalty in the recoverable cell −1.0 point instead of 0, which puts $\Delta_{H1}$ 1 point above 100. One refused single prompt makes $\Delta_{H3}$ −0.25 instead of 0. Neither changes a reading.

- **Cost.** The window again cost more in dollars while sending fewer tokens: 3.3 to 3.7 times full history's billed cost per episode, at 0.84 to 0.90 times the tokens, because dropped turns break prompt caching and this provider bills the rewritten prompt as cache writes at 1.25 times the input price. These ratios depend on each provider's prices and tokenizer and are not compared across models.

With one model per provider, the agreement cannot be attributed to any one factor, and it says nothing about other models, other providers or real tasks. These readings come from a committed script that reads the counted records (`analysis/replication_readings.py`).

## 5 Decomposing ARC-AGI-3's harness gap

**Configurations.** We run Gemini 3.8 Flash (Google 2026a) at its high thinking level in four configurations of ARC-AGI-3's benchmarking harness (upstream commit 4ca00e6 of 2026-09-21, plus our patches: 18 in all, the last of which applies only to the runs under amendment R7):

- **S0**, the Standard harness: earlier turns are sent back as visible text only (observations and replies, no reasoning items), its system prompt carries the note-taking instruction, and when the context passes 175k characters the oldest turns are dropped;
- **S1**: as S0, but old turns are never dropped: once a request reaches 175k provider tokens, the harness replaces the history with a model-written summary (a later trigger than S0's 175k characters, which we count as part of the loss policy);
- **S2**: S1 without the note-taking instruction;
- **S4**, the Provider Adapter: native replay of the model's reasoning items, the same summary trigger (its summary request also carries the replayed reasoning), and no note-taking instruction.

Two known differences from the benchmark's published Standard runs: all our configurations request thought summaries (logged, and never sent back in S0 to S2), which the published Standard recordings do not carry, and our patches add a transport-level retry that the published runs did not have. Our S0 is therefore close to, not identical with, the published Standard harness.

Each step of the path S0 → S1 → S2 → S4 is designed to change one component: S0→S1 the loss policy, S1→S2 removes the notes instruction, and S2→S4 adds native replay with its item format (the two cannot be separated here). The Adapter has no notes instruction, so the instruction has to come out somewhere along the path. Removing it before replay is added keeps the last step to replay. Amendment R7 (§8) raised the S1 and S2 summary output budget from 8,192 to 65,536 tokens partway through, so S2→S4 can also lower it (S4 keeps 8,192, which never bound in any S4 run: 132 summaries, at most 3,073 output tokens), and in three games the S1→S2 step also changes it (see "How firm this is"). A path through a configuration with both notes and replay would add a component at every step, and we did not run one. Each share is a step's change in mean score divided by the gap $G = \bar{R}_{S4} - \bar{R}_{S0}$, so $\sigma_{loss}$ (S0→S1), $\sigma_{notes}$ (S1→S2) and $\sigma_{replay}$ (S2→S4) sum to 1, where $\bar{R}$ is the mean over games of ARC's per-game score, RHAE (Relative Human Action Efficiency) (ARC Prize Foundation 2026a): a score from 0 to 100 that credits each cleared level by how its action count compares with a human baseline, with later levels weighted more.

**Games and runs.** Four games (FT09, SB26, R11L, CD82), fixed before any run by a registered rule rather than drawn: they are the only public games whose two published high-thinking runs each cost at most $60, whose published Standard-to-Adapter gap is at least 20 points, and whose published Standard run ended normally (ARC Prize Foundation 2026d). One counted run per configuration and game, under the benchmark's 12-hour clock and per-level action budgets.

**Results.** Under the registered rules, a run that ended by an API or agent error was re-run once, and a run that hit the 12-hour clock was re-run if its model calls had been more than twice as slow as the registered yardstick. The re-run counts, and a re-run that ends the same way counts as it stands. Four of the 16 counted runs are not their unit's first run: FT09 S0 (a latency re-run), R11L S2 and CD82 S1 (error re-runs, where CD82 S1's re-run also ended on an error and counts as is) and FT09 S2 (a replacement under amendment R7). Table 2 and Figure 2 give each counted run's score, and Appendix A lists all 21 runs with their endings.

Table 2: ARC-AGI-3 score (RHAE) of each counted run, by game and configuration.

| Game | S0 | S1 | S2 | S4 |
|---|---|---|---|---|
| FT09 | 18.11 | 100.00 | 18.06 | 100.00 |
| SB26 | 5.44 | 100.00 | 100.00 | 100.00 |
| R11L | 6.45 | 11.21 | 14.29 | 14.29 |
| CD82 | 6.94 | 41.11 | 12.95 | 73.02 |
| **Mean** | **9.24** | **63.08** | **36.32** | **71.83** |

The gap is G = 62.59 (90% bootstrap interval over games [29.51, 87.44]). Its shares are $\sigma_{loss} = 0.860$ [0.541, 1.000], $\sigma_{notes} = -0.427$ [−0.879, 0.011] and $\sigma_{replay} = 0.567$ [0.000, 0.960]. With four games there are only 35 distinct bootstrap resamples, so these intervals are coarse. No resample gave $G \leq 0$. We registered two competing predictions: A, that the provider-neutral loss policy carries a large share, replay at most half and notes almost none ($\sigma_{loss} \geq 1/3$, $\sigma_{replay} \leq 1/2$, $|\sigma_{notes}| \leq 0.10$), and B, that native replay carries at least half ($\sigma_{replay} \geq 1/2$). On the point estimates, under amendment R7 (§8), B holds and A does not. Under the rules as frozen before R7, neither holds (below). $\sigma_{replay}$'s interval lies on both sides of B's threshold.

The path is not monotone. Summary compaction with the note-taking instruction (S1) reaches 63 of the Adapter's 72 points on these games without native replay. Removing the instruction (S2) loses 27 points, and replay (S4) recovers them and more. Because the shares sum to one, the negative $\sigma_{notes}$ inflates the other two.

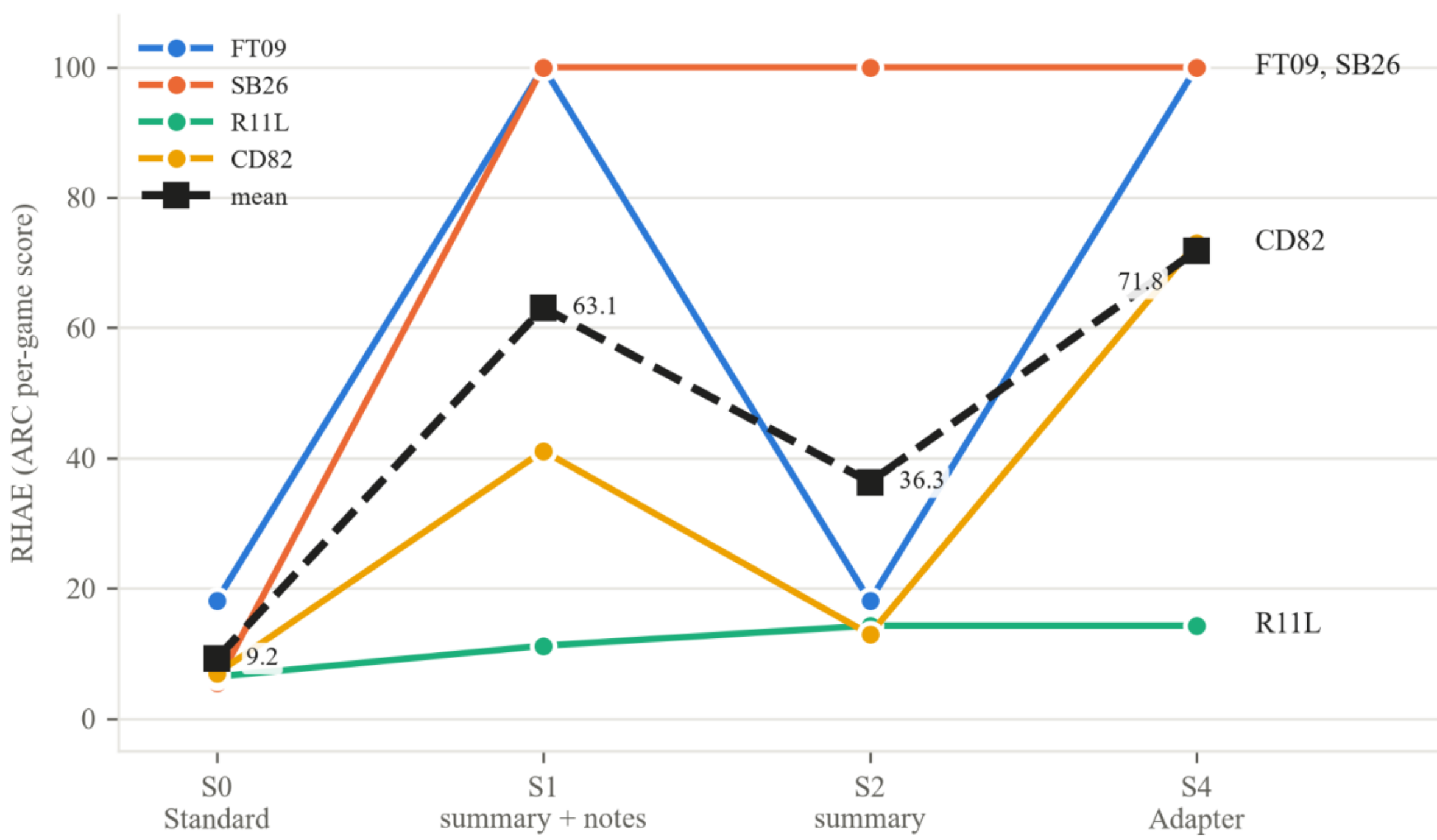

Figure 2: ARC-AGI-3 score by harness configuration, one line per game. The dashed line is the mean.

**How firm this is.** Not very, and we say so:

- *Prediction B rests on one replaced run.* Under the rules as frozen, FT09 S2's counted run was its error re-run, which ended when its summary requests came back empty, after clearing 5 of 6 levels (71.43). Amendment R7, adopted after that run's levels had been seen, raised the summary budget and gave the unit one replacement run, which cleared 3 (18.06). That replacement is the FT09 S2 score in Table 2. Counting the earlier run instead changes one of the 16 scores, from 18.06 to 71.43, and so S2's mean from 36.32 to 49.67. The S1 and S4 means and the gap are unchanged, but S2 then recovers 65% of the gap instead of 43%, the replay step's share falls from 57% to 35% ($\sigma_{replay} = 0.354$), and neither prediction holds. The headline replay share therefore hinges on which of FT09 S2's two runs counts. Leaving FT09 out altogether gives $\sigma_{replay}$ = 0.357, and leaving CD82 out gives 0.445, so B also fails without CD82.

- *Runs vary.* FT09 S0's first run hit the 12-hour clock with calls more than twice as slow as in the pilot, so the registered latency rule re-ran it. The re-run was just as slow (median 176 s per call against 181 s) and also hit the clock, yet the two runs scored 58.17 and 18.11, 40 points apart. One run per cell cannot separate a configuration's effect from this variation.
- *Mixed budgets.* Because R7 changed the summary budget partway through, FT09's and SB26's S1 runs played under 8,192 tokens and their S2 runs under 65,536 (R11L the reverse), so in those games the S1→S2 step behind $\sigma_{notes}$ also changes the budget.
- *Time.* The configurations ran on different days, against a model with no dated snapshot.
- *Selection.* The games were chosen for a large published gap, so the shares describe games where the Adapter is known to help.

Scoring each unit's original run instead of its re-run gives G = 52.57 and $\sigma_{replay}$ = 0.671. $\sigma_{replay}$ reaches the registered 2/3 bar for a replay-dominated gap in that reading and at 150 actions (below), but not on the primary reading. S0's model calls were much slower than S4's (median 51 to 176 s per call against 9 to 32 s), and two of its counted runs ended on the 12-hour clock. Scoring every run on its first 150 actions, a secondary reading added after S0's first run hit the clock (amendment R5), gives a gap of 53.95 and $\sigma_{replay}$ 0.711 (0.464 with FT09 S2's replaced run back). No counted run was censored at 150 actions.

## 6 Clear rates by level depth

H5 asks whether the Adapter's advantage in clearing a level, once it is reached, is larger at deep levels (4 and beyond) than at early ones (1–2). Level 3 is shown but is not in the contrast.

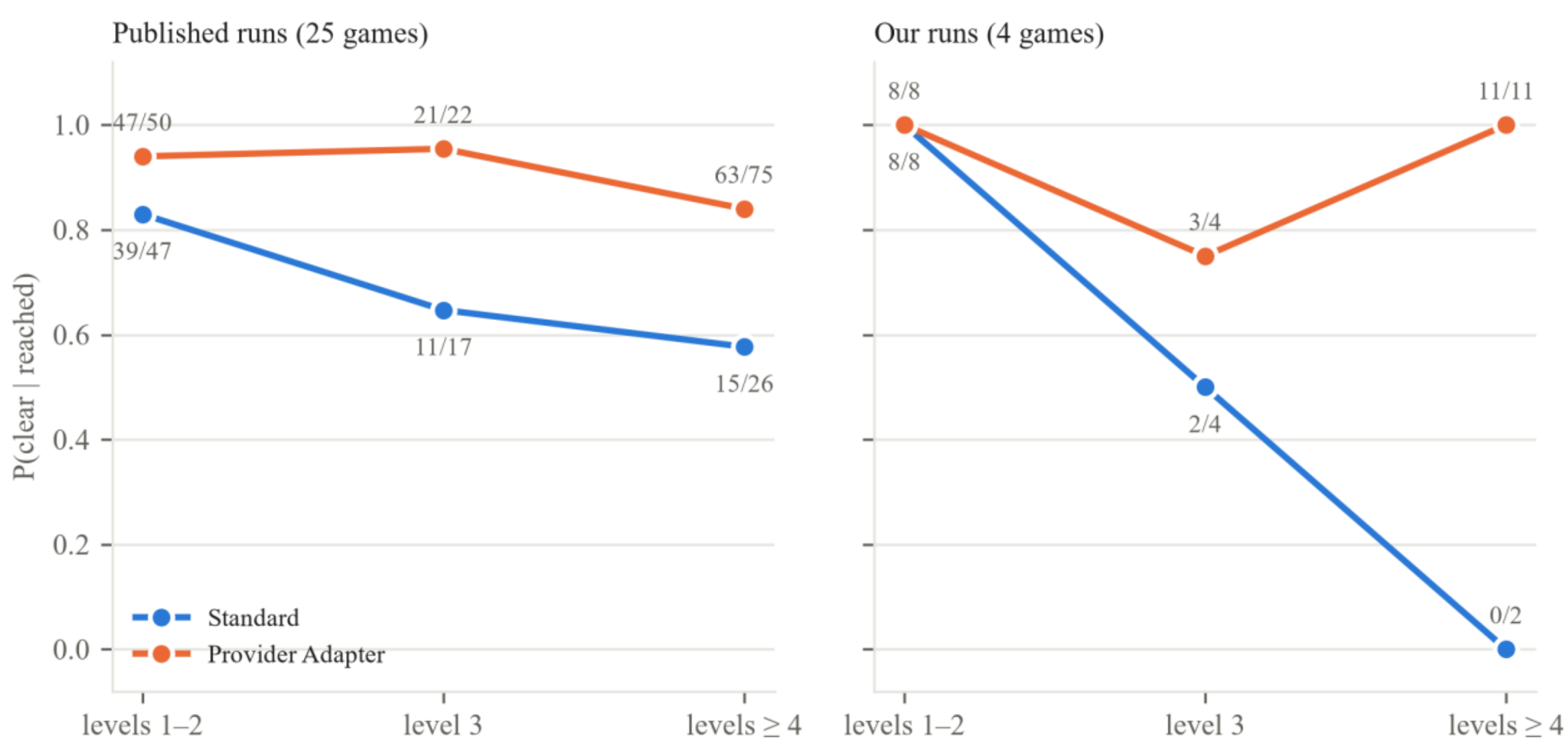

Figure 3: P(clear | reached) by level band, Standard against Provider Adapter. Left: the benchmark's published runs of the same model at high thinking (25 games). Right: our runs, S0 against S4 (4 games). Labels give cleared/reached. Level 3 is shown but is not in the registered contrast.

**Published runs.** In the benchmark's published runs of the same model at high thinking on the 25 public games (ARC Prize Foundation 2026d), the Standard harness cleared 39 of 47 early levels it reached and 15 of 26 deep ones. The Adapter cleared 47 of 50 and 63 of 75. In a logistic model with game fixed effects, the Adapter's advantage in the probability of clearing is +0.14 early and +0.44 deep, a difference of C = +0.30 (90% interval [0.16, 0.46], or +0.26 under Firth's penalized fit). Dropping the levels at which a run ended by an inferred error or a closed session gives C = +0.22 ([0.05, 0.41]).

**Our runs.** On our four games both harnesses cleared every early level they reached. At deep levels S0 cleared 0 of 2 and S4 11 of 11 (C = +0.75 [0.67, 1.00], Figure 3).

**What this does and does not show.** Both readings point the same way: the harnesses differ much more late in a game than at its start. But levels are played in order, so each run has at most one level it reached and did not clear, the one it stopped on, and these rates largely restate where runs stopped and which runs won (the Adapter won 9 of the 25 published games, the Standard harness none). On the logit scale the published difference is not distinguishable from zero, so the result does not separate a growing advantage from a constant one compressed by near-ceiling early clear rates. The published runs were in effect seen before the analysis was written (their per-level endings were tabulated to plan the study), and they are the runs whose gap motivated it. Our own runs' scores were also known before this analysis was computed.

## 7 Exploratory: compaction and unusable replies

Summary-based configurations compact often: once every 6.3 steps in S1, 6.9 in S2 and 5.6 in S4. The model sometimes returned a reply with no usable action: empty text, or a call to a tool the harness never offered. This happened 22 times in S1 and 17 in S2, and never with native replay (S4, 737 steps). In S1 such replies were more frequent in the five steps after a compaction (3.4% of calls) than elsewhere (1.3%). S2 showed no such pattern, and level clears per step did not differ consistently. Time since a compaction is confounded with time into the run, and the counts are small. The analysis was specified after most endings had been seen, prompted by CD82 S1's two failures, in which the model returned four unusable replies in a row soon after a compaction. Those failures are among the counts above and are not independent support. We did not count unusable replies in S0, which never compacts, so this does not separate compaction from the visible-text format that S0 to S2 share. We report it as a lead, not a finding.

## 8 Deviations from the pre-registration

Table 3 lists every change to the registered plan, each with the dated record that made it: an amendment to the pre-registration (R1 to R8) or an entry in the decisions log.

Table 3: Changes to the pre-registered plan.

| Change | Record | Effect on this paper |
|---|---|---|
| **Minimal path after a cost checkpoint,** decided after the S0 and S4 runs' results were known. Dropped: a second run of every ARC configuration and game, the registered phase 2 (a state component evaluated on three held-out games), the replication of H1 on a second model from the same provider, an open-weights run, the hand-labelling of failures and a planned redesign pass. The registered three-game fallback was not used. Also: the probe's template count was capped at 200 (the registered power rule would have set 290), S1 and S2 ran after S0 and S4 rather than in one wave, H5 on our runs became an analysis of seen data, and the latency yardstick for a configuration's first run was scaled by S0's measured slowdown. | Amendment R6 (2026-10-09) | One counted run per ARC configuration and game, H1's confirmatory test on one model (R8 later added a replication, below), no state component evaluated and power below target (§3). The yardstick change altered nothing: no S1 or S2 run ended on the clock. |

| Change | Record | Effect on this paper |
|---|---|---|
| **H2 stopped.** H2 tested whether replaying the model's reasoning carries derived facts, on a second model from the same provider. Two registered engagement checks found too little reasoning at the steps that showed the clues for replay to carry anything (0 of 20 calls, then 6 of 20 on a harder variant, against a bar of 10), so H2 stopped. | Amendments R2 and R3 (2026-10-02), with the stop recorded 2026-10-03 | Reported as not testable in this task family. The probe does not test reasoning replay. That model's H1 replication met its validity-gate thresholds (80 of 80 restated, 9 of 80 withheld), but its main runs were dropped (R6). |
| **Holm family.** With H2 stopped, the correction covers H1 and H3. | Decision of 2026-10-06, before any H1 or H3 test, registered in amendment R6 | Over the registered family of four, with H2's tests counted as not rejected, H1 and H3 are still supported. |
| **Action-matched readings.** H4's quantities are also computed at a matched 150 actions, added after S0's first run hit the 12-hour clock. | Amendment R5 (2026-10-07) | Secondary readings that no decision rule reads (§5). No counted run was censored. |
| **Summary output budget.** Three S2 runs (FT09's run and its re-run, R11L's first run) ended because the summary request came back empty four times in a row: at high thinking, the model's reasoning used up the request's 8,192-token output budget. The S1 and S2 summary budget was raised to 65,536 tokens, the budget of every action request, and FT09 S2 got one replacement run. | Amendment R7 (2026-10-10) | Adopted after the replaced runs' endings and levels had been seen. Under the rules as frozen, FT09 S2 counts at 71.43, $\sigma_{replay}$ is 0.354 and neither prediction holds (§5). R11L S2's re-run had already ended normally under the old budget and counts. Three of the eight counted S1 and S2 runs (FT09 S1, SB26 S1, R11L S2) played under the old budget. S4 keeps 8,192, which never bound. The upstream harness later removed the 8,192-token cap for the same failure (Mazur 2026). |
| **A missed run.** SB26 S2 was left out of R7's run list by error. | Decisions log, 2026-10-12 | It ran after the others. |
| **Re-run timing.** FT09 S0's re-run, which the registered rules start right after the run it replaces, ran last in its block. | Decisions log, 2026-10-07 | The re-run counts, as registered. Only its place in the run order differs. |
| **Analysis rulings.** H1's model check uses Firth's penalized fit for separated cells, ruled before any sweep result was read. H5 uses the maximum-likelihood limit where the contrast is identified, else Firth, ruled before any H5 quantity was computed but after our S0 and S4 runs' outcomes and the published runs' endings were known. | Decisions log, 2026-10-10 (two entries) | Fixes how the model checks read cells with no variation. On our runs, Firth decides H5's contrast (§6). |
| **Analysis fixes made after the results were read.** The Firth fit's step cap rose from 100 to 1,000, and H3's model check was allowed to fit one template. | Decisions log, 2026-10-10 (two entries) | H1 and H3 first read "inconclusive". Both labels are reported (§4). |

| Change | Record | Effect on this paper |
|---|---|---|
| **Pilot blinding.** One ARC pilot run's level count was displayed during the pilots. | Decisions log, 2026-10-02 | Pilot runs enter no analysis, and no open design choice depended on it. |
| **A transport re-run.** One of gpt-6-luna's PersistGym-lite episodes (of 9,680) failed when the provider's Flex tier was temporarily unavailable (HTTP 429). | Pre-registration §13 and decisions log, 2026-10-10 | Re-run once, as registered. |
| **A replication on a second provider's model.** Amendment R8 registered H1 and H3 on claude-haiku-5-5. It was written with gpt-6-luna's results known and before any output from this model existed. Its registered cost rule dropped W-n-0. The cost forecast priced full history from the validity run rather than the pilot, with the same outcome. Its validity runs could not estimate two power assumptions (a template spread and a no-memory success rate), so substitutes were chosen before any result was read. Two runs stopped on the provider's rate limit and were continued, with three transport re-runs, at lower parallelism. Before any main episode, the harness's token count for whitespace-only replies was aligned with the amendment, and no reply was affected. | Amendment R8 (2026-10-18) and decisions log, 2026-10-18 and 2026-10-19 | Reported as a replication outside the Holm family, with no confirmatory label (§4). The notes cell has no counterpart on this model, and its power rests on the substitutes (§3). |
| **Other amendments.** R1 to R4 concern a pilot probe and the same-provider second model's arm. R2 also added a rule excluding any episode in which a response's echoed reasoning effort differs from the request. | Amendments R1 to R4 (2026-10-02 to 2026-10-03) | That arm is not reported, and the exclusion rule excluded no episode. |

## 9 Limitations

- **One model per provider.** The probe's confirmatory test uses one model, and its replication one model from a second provider. The two differ at once in provider, training, reasoning, API and serving path, so their agreement cannot be credited to any one of these, and it says nothing about other models, other providers or real tasks. ARC uses one model and one counted run per configuration and game. The ARC shares are point estimates, and run-to-run variation is of the same order as some effects.
- **Replay is not isolated:** the probe does not test replay, because H2 stopped (§8), and the ARC replay step bundles replay with the provider's item format and, in some games, a change of summary budget.
- **Our S0 approximates the published Standard harness** (thought summaries requested, throttled calls re-tried).
- **Selected games:** the four ARC games were chosen for a large published gap.
- **Synthetic probe:** PersistGym-lite results are not claimed to transfer to other agents or tasks. The window's success penalty when the fact is lost is expected by construction and only confirms that the window removed the fact (a registered manipulation check). The finding is the contrast with the recoverable case.
- **Time:** ARC configurations ran on different days, against a model without a dated snapshot.

- **Costs are provider- and period-specific:** the window's dollar ratios follow each model's cache pricing and the prices of 2026-09-30 and 2026-10-17.
- **Not a model comparison:** nothing here ranks models or providers.

## 10 Reproducibility and cost

All code, the pre-registration (`PREREG.md`) with its dated amendments, the decisions log, the per-run metadata and the cost ledger are in the repository at https://github.com/raimondasl/persistgym. Every number in this paper is produced by a committed script and was re-derived independently before posting. Most scripts read only committed run outputs. The carried-fact accounting in §4 and the compaction analysis in §7 also read the raw model transcripts, which are kept out of the repository. Model calls cost $750.11 in all (billed estimate), split by the approval blocks of the cost ledger: pilots and checks $38.71, the ARC path runs $507.96 ($290.22 for S0 and S4, $217.74 for S1 and S2), the PersistGym-lite sweep $73.75 and its replication $129.69.

The code, analyses, reviews and manuscript drafts were produced with an AI coding assistant (Claude Code) under the author's direction. The author designed the study, made every recorded decision, and is responsible for all text, code and results. This work reflects the author's own views, not those of Microsoft.

## Appendix A: every ARC run

Table 4: Every ARC run, by game and configuration, in the order run. "Agent error" covers the summary failures and the unusable-reply endings described in §5, §7 and Table 3.

| Game | Config | Run | Ending | Levels | Actions | RHAE | Status |
|---|---|---|---|---|---|---|---|
| FT09 | S0 | 1 | 12-hour clock | 5 of 6 | 236 | 58.17 | Re-run under the latency rule |
| FT09 | S0 | 2 | 12-hour clock | 3 of 6 | 214 | 18.11 | Counted |
| FT09 | S1 | 1 | Won | 6 of 6 | 92 | 100.00 | Counted |
| FT09 | S2 | 1 | Agent error | 3 of 6 | 49 | 28.57 | Re-run after an error |
| FT09 | S2 | 2 | Agent error | 5 of 6 | 112 | 71.43 | Counted under the frozen rules, replaced under R7 |

| Game | Config | Run | Ending | Level s | Actions | RHAE | Status |
|---|---|---|---|---|---|---|---|
| FT09 | S2 | 3 | Action budget | 3 of 6 | 219 | 18.06 | Counted (R7 replacement) |
| FT09 | S4 | 1 | Won | 6 of 6 | 112 | 100.00 | Counted |
| SB26 | S0 | 1 | Action budget | 3 of 8 | 238 | 5.44 | Counted |
| SB26 | S1 | 1 | Won | 8 of 8 | 131 | 100.00 | Counted |
| SB26 | S2 | 1 | Won | 8 of 8 | 175 | 100.00 | Counted |
| SB26 | S4 | 1 | Won | 8 of 8 | 138 | 100.00 | Counted |
| R11L | S0 | 1 | Action budget | 2 of 6 | 365 | 6.45 | Counted |
| R11L | S1 | 1 | Action budget | 2 of 6 | 325 | 11.21 | Counted |
| R11L | S2 | 1 | Agent error | 1 of 6 | 28 | 4.76 | Re-run after an error |
| R11L | S2 | 2 | Action budget | 2 of 6 | 286 | 14.29 | Counted |
| R11L | S4 | 1 | Action budget | 2 of 6 | 281 | 14.29 | Counted |
| CD82 | S0 | 1 | 12-hour clock | 2 of 6 | 235 | 6.94 | Counted |
| CD82 | S1 | 1 | Agent error | 2 of 6 | 87 | 14.29 | Re-run after an error |
| CD82 | S1 | 2 | Agent error | 4 of 6 | 210 | 41.11 | Counted as it stands (a second error) |
| CD82 | S2 | 1 | Action budget | 2 of 6 | 297 | 12.95 | Counted |
| CD82 | S4 | 1 | Won | 6 of 6 | 206 | 73.02 | Counted |